\documentclass{article}

    \usepackage[preprint]{neurips_2020}

\usepackage[utf8]{inputenc} % allow utf-8 input
\usepackage[T1]{fontenc}    % use 8-bit T1 fonts
\usepackage{hyperref}       % hyperlinks
\usepackage{url}            % simple URL typesetting
\usepackage{booktabs}       % professional-quality tables
\usepackage{amsfonts}       % blackboard math symbols
\usepackage{nicefrac}       % compact symbols for 1/2, etc.
\usepackage{microtype}      % microtypography

\usepackage{amsmath}

\usepackage{graphicx}

\title{Band-limited Soft Actor Critic Model}

\author{%
  Miguel Campo\thanks{Equal contribution. Correspondence to miguelca@fb.com} \qquad Zhengxing Chen \qquad Luke Kung \qquad Kittipat Virochsiri \qquad Jianyu Wang \\
   \\
  Facebook\\
  \{miguelca,czxttkl,lkung,kittipat,jianyu\}@fb.com \\
}

\begin{document}

\maketitle

\begin{abstract}
  Soft Actor Critic (SAC) algorithms show remarkable performance in complex simulated environments. A key element of SAC networks is entropy regularization, which prevents the SAC actor from optimizing against fine grained features, oftentimes transient, of the state-action value function. This results in better sample efficiency during early training. We take this idea one step further by artificially bandlimiting the target critic spatial resolution through the addition of a convolutional filter. We derive the closed form solution in the linear case and show that bandlimiting reduces the interdependency between the low and high frequency components of the state-action value approximation, allowing the critic to learn faster. In experiments, the bandlimited SAC outperformed the classic twin-critic SAC in a number of Gym environments, and displayed more stability in returns. We derive novel insights about SAC by adding a stochastic noise disturbance, a technique that is increasingly being used to learn robust policies that transfer well to the real world counterparts.

\end{abstract}

\section{Introduction}
\label{submission}

Deep reinforcement learning has become an important tool for robotics tasks in high dimensional environments. Soft Actor-Critic (SAC) [1] has emerged as a robust candidate that incentivizes exploration in the continuous action space as a mechanism to avoid suboptimal convergence, e.g., convergence to local maxima. SAC is a model free, off-policy, sample efficiency algorithm. Those traits make it well suited for environments where simulated training doesn't transfer well to real world situations, like object pushing, and environments where realistic simulated training isn't unavailable, like product recommendation systems.

SAC algorithms' superior performance is normally attributed to sample efficient policy exploration in the action space. An aspect of SAC that hasn't received much attention, and that can help understand its superior performance, is the trade off between policy exploration and spatial resolution. Spatial resolution refers, broadly speaking, to the distance between distinct features of the state-action value function. During the policy \emph{improvement} step (actor network), Monte Carlo sampling in the action space averages out rapid variations of the state-action value function; that is, filters out high frequency components. During the policy \emph{evaluation} step, the critic network prioritizes the learning of slow variations (low frequency components) of the state-action value function. This is known as the \emph{frequency principle} or the \emph{spectral bias} of deep neural networks [5], and it helps guide early policy exploration towards smooth value maxima, for which Monte Carlo sampling during policy improvement is well suited. Spectral bias provides another interpretation of the high returns and superior convergence properties of SAC algorithms.

The asymmetric effect of spectral bias on the actor and critic networks is not well understood though. Monte Carlo sampling in the action space prevents the actor from learning policies that exploit high frequency spatial components, like narrow maxima, but doesn't prevent the critic from \emph{eventually} learning those components. This is important because high frequency components are harder to learn and may have higher variance [19], and learning them can 'pass' the variance to the low frequency components which, in classic SAC, are strongly coupled with the high frequency components. Strong coupling and transfer of variance to the low frequency components may hurt performance even in simple environments because the low frequency components are responsible of the location of the value maxima. Added variance in the late training stages may cause policies to drift.

One way to limit this effect is to \emph{explicitly} constrain the ability of the critic to learn the high frequency components of the state-action value function, a process that is called bandlimiting (see [6] for an application of bandlimiting in supervised tasks). We hypothesize that bandlimiting could have a positive impact on convergence because it would incentive the critic to learn the most relevant low frequency components. In this work we simulate the effect of adding a very simple (separable) convolutional filter to \emph{bandlimit} the state-action value approximation during policy evaluation. In order to avoid overconstraining the critic network, we match the filter bandwidth to the policy bandwidth, which itself varies as a result of policy improvements. That is, we constrain the ability of the critic to learn fine grained details of the value function, but only those details \emph{that are closer to each other than the ability of the actor at each step to discern them}. 

We test this approach in a number of Gym environments and show that bandlimiting may lead to asymptotic improvements in performance and, importantly, in stability. We then add noise to the environment to illustrate how the bandlimiting filter reduces the influence of high frequency components and improves returns. Noise analysis is becoming increasingly important in reinforcement learning. Although added noise erodes performance during simulations, it's known to help learn policies that generalize to non ideal, real world tasks [9]. Overall, the results point to the effect of bandlimiting as a way to reduce the influence of high frequency components on the state-action value approximation and to guide the critic to learn features that are useful for the actor.

\section{Related Work}

\subsection{Frequency principle}

A relatively new area of research that is related to our work studies the behavior of deep neural networks in the frequency domain--that is, how efficient the network is at learning the different frequency components of the target function. Recent results have shown that a deep neural network tends to learn a target function from low to high frequencies during the training [5], and that networks trained with a bandlimited spectra version of the data--a version of the data in which high frequency components have been eliminated--are more efficient at learning to use the low frequency components [6]. These results could translate to reinforcement learning algorithms and, specifically, to actor critic models. We would expect the critic network to learn first the low frequency components of the state-action value approximation, and we would expect that a bandlimited critic would be more efficient at learning the low frequency components of the approximation, thus avoiding early suboptimal decisions that are known to slow down training [18]. Along these lines, Feng et al. [13] have recently proposed a loss function to minimize the Bellman residual error that implicitly prioritizes the low frequency components of the Bellman error. In this study we build on these ideas to explicitly bandlimit the output of the target critic, which plays the role of \emph{supervised} label in actor-critic algorithms.

\subsection{Target policy smoothing}

Bandlimiting the target critic is a form of regularization that resembles Q function averaging, which is used in deterministic RL algorithms to help reduce the variance of the value estimates. Q function averaging is based on the idea that proximal actions should have similar values. Fujimoto et al. [11] have used this regularization strategy for deep value learning and they refer to it as target policy smoothing, which is itself based on the original SARSA algorithm [15]. In their proposed algorithm, the authors average the value of the target critic by bootstraping off of value estimates associated to small deviation in the chosen action. Nachum et al. [2] use a similar idea but instead of smoothing out the value of the target critic they smooth out the value of the critic itself. Our approach is closer to [11] because we also bootstrap off of value estimates associated to small deviation in the chosen action. In our case, though, the deviations and the weight applied to the value estimates are chosen in order to achieve strong attenuation of the high frequency components of the state-action value approximation.

\subsection{Representation learning for policy and value}

In complex environments there could be a mismatch between the ability of the critic to learn complex value functions and the ability of the actor to exploit them. Recent research in reinforcement learning seeks to close this gap by improving the ability of the actor to fully exploit multimodal state-action value functions. Monomodal assumptions can lead to policies with undesirable policies, like policies that don't explore enough, or policies that exploit actions that are located in \emph{valleys of the value function} (a phenomenon that resembles \emph{aliasing} [20] and that we've noticed in our own recommendation systems where we have observed that SAC algorithms can converge to policies that awkwardly recommend low value products that are themselves located between peaks of high value products). One successful approach to avoid suboptimal policies is to learn complex policy distributions that depart from the Gaussian monomodal parametrization (see, e.g., [4]), thus reducing the mismatch between the properties of the policy parametrization and the properties of the state-action value approximation. In our analysis we also close this mismatch, but we do it by matching the spatial resolution of the two functions.

\section{Markov Decision Process}

The standard SAC algorithm assumes an infinite-horizon Markov decision process (MDP) described by an action space $\mathcal{A}$, state space $\mathcal{S}$, a reward function $r$: $\mathcal{S} \times \mathcal{A} \rightarrow \mathbb{R}$, and a transition function $P$: $\mathcal{S} \times \mathcal{A} \times \mathcal{S} \rightarrow \mathbb{R}$. The agent interacts with the MDP and selects an action $a_t \in \mathcal{A}$ based on the current state $s_t \in \mathcal{S}$. After the action is selected, the agent will transitioning to a new state $s_{t+1} \sim P(s_t, a_t)$ and receive a reward $r_t = r(s_t, a_t)$. The agent assumes a discount factor $\gamma \in [0, 1)$ and its choices are governed by a stochastic policy such that $a_t \sim \pi(\cdot|s_t)$. The agent will chose a policy to maximize the expected sum of discounted returns, which defines the value function:
\begin{equation} \label{value}
V^{\pi}(s) = E_{\pi} \left[ \sum \gamma^{t} r_{t} | s_{0} \right]
\end{equation}
Similarly, the sum of expected discounted rewards from taking action \(a_t\) in state \(s_t\) defines the state-action value function (or Q-function) which can be written as
\begin{equation} \label{value}
Q^{\pi}(s_{t},a_{t}) = E_{\pi} \left[ \sum_{i=t} \gamma^{i-t} r_{s_{i},a_{i}} | s_{t},a_{t} \right]
\end{equation}

\subsection{Soft Actor Critic}

In the SAC algorithm, we update the policy towards the Boltzmann policy with temperature \( \alpha \), with the Q-function taking the role of (negative) energy. Practically speaking, we update the policy by minimizing the Kullback-Leibler divergence between the policy and the Boltzman policy,
\begin{equation} \label{pinew}
\pi_{new} = arg min_{\pi'} D_{KL} \left( \pi'(\cdot|\boldsymbol{s}_t) || \frac{exp( \frac{1}{\alpha} Q^{\pi_{old}(\boldsymbol{s}_t,\cdot)})}{
Z^{\pi_{old}(\boldsymbol{s}_t)}
} \right)
\end{equation}
In the classic SAC formulation, minimizing the expected KL-divergence to learn the policy parameters is equivalent to maximizing the expected value of the state-action value function plus a regularization term that captures the entropy of the distribution of possible actions,
\begin{equation} \label{J_base}
J_{\pi}(\phi) = \alpha E_{\boldsymbol{s}_t\sim D,\boldsymbol{\epsilon}_{t}\sim N} \left[
log \pi_{\phi}(f_{\phi}(\boldsymbol{\epsilon}_{t},s_t)|\boldsymbol{s}_t) 
\right]
 - E_{\boldsymbol{s}_{t}\sim D,\boldsymbol{\epsilon}_{t}\sim N} \left[
Q_{\theta}(\boldsymbol{s}_t,f_{\phi}(\boldsymbol{\epsilon}_{t},\boldsymbol{s}_t))
\right]
\end{equation}
where 
\begin{equation} \label{action}
\boldsymbol{a}_t = f_{\phi}(\boldsymbol{\epsilon}_t; \boldsymbol{s}_t)
\end{equation}
is the stochastic action that depends on the state and on a stochastic term \(\epsilon_t\). The classic policy reparametrization most frequently used in practice is
\begin{equation} \label{popularreparam}
f_{\phi}(\boldsymbol{\epsilon}_t; \boldsymbol{s}_t) = tanh \left( \mu_{\phi}(\boldsymbol{s}_t) + \sigma_{\phi}(\boldsymbol{s}_t) \boldsymbol{\epsilon}_t \right)
\end{equation} 
with \(\epsilon_t\) a r.v. that has spherical Gaussian distribution and \(\mu_{\phi}(\boldsymbol{s}_t)\) and \(\sigma_{\phi}(\boldsymbol{s}_t)\) the mean column vector and diagonal covariance matrix, respectively.

\section{Averaged state-action values}

Exploration helps train robust agents in complex environments. In SAC algorithms, exploration is achieved by Monte Carlo sampling the actions from a parametrized stochastic policy. Sampling increases the range of actions and states visited. Sampling also limits the spatial resolution of the policy by averaging out the features of the state-action value function.

To see this, it's convenient to reinterpret the state-action value expectation term in eq. (\ref{J_base}) as a convolution between the Q function and the Gaussian policy kernel. The practical effect of the convolution is to smooth out the peaks and valleys of the Q function in the action space, for the purposes of the policy update. If two distinct peaks of the Q function are too close to each other, they get blurred together into one. 'Too close' means that they fall within the support of the policy kernel. 

The loss of spatial resolution of the state-action value function in the action space is an intrinsic feature of SAC algorithms and is determined by the variance of the policy distribution. Reducing the weight of the entropy regularization term reduces the variance of the policy and enables the actor to learn the sharp location and value of the different peaks of the state-action value function. But this can also degrade the performance of SAC algorithms by curtailing exploration, which is key to find global value maxima, and by increasing the sensitivity to hyperparameter values.

\section{Bandlimited value}

Entropy regularization prevents the SAC actor from optimizing against fine grained features of the state-action value function, and improves sample efficiency. Explicitly limiting the spatial resolution of the critic could be a way to further increase sample efficiency. To see this, we rewrite the state-action value expectation in (\ref{J_base}) as a convolution, 
\begin{equation} \label{main_becomesconv}
E_{\boldsymbol{s}_{t}\sim D,\boldsymbol{\epsilon}_{t}\sim N} \left(Q_{\theta}(\boldsymbol{s}_t,f_{\phi}(\boldsymbol{\epsilon}_{t},\boldsymbol{s}_t))
\right)
= E_{\boldsymbol{s}_{t}\sim D}  
\left[
\left( 
Q_{\theta}(\boldsymbol{s}_t, .) *  
f_N( .;\sigma_{\phi}(\boldsymbol{s}_t) )
\right)
_{\mu_{\phi}(\boldsymbol{s}_t)}
\right]
\end{equation}
where * is the convolution operator, evaluated at the policy mean \(\mu_{\phi}(\boldsymbol{s}_t)\), and \(f_N\) is the Gaussian probability distribution (see details of this derivation in appendix B). Because of the properties of the convolution in the frequency domain, equation (\ref{main_becomesconv}) implies that the high spatial frequency components of the state-action value function have little or no impact on the expectation. In effect, the Gaussian kernel acts as a filter that eliminates any high frequency components of the state-action value approximation that the critic might have learnt. Importantly, this effect is more pronounced for high temperature values associated with high exploration. It can be seen that there is an inverse relationship between policy exploration levels and the selectivity of the Gaussian filter. This filter dampens or eliminates the frequency components of the state-action value approximation above a certain threshold value $w_{cutoff}$ which is given by the inverse of the policy standard deviation:
\begin{equation} \label{main_cutoff}
w_{cutoff}^{i} \approx \frac{\pi}{2\sigma_{\phi}^{i}(\boldsymbol{s}_t)}
\end{equation} 
With this in mind, we can rewrite the state-action value expectation as
\begin{equation} \label{main_approxQexpectation}
E_{\boldsymbol{s}_{t}\sim D,\boldsymbol{\epsilon}_{t}\sim N} \left(Q_{\theta}(\boldsymbol{s}_t,f_{\phi}(\boldsymbol{\epsilon}_{t},\boldsymbol{s}_t))
\right) 
\approx
E_{\boldsymbol{s}_{t}\sim D} \left(Q_{\theta}^{LOW}(\boldsymbol{s}_t,f_{\phi}(\boldsymbol{\epsilon}_{t},\boldsymbol{s}_t))
\right)
\end{equation}
where, in eq. (\ref{main_approxQexpectation}), \(Q_{\theta}^{LOW}\) refers to the the function that results from removing the high frequency components from the state-action value approximation \(Q_{\theta}\). Eq. (\ref{main_approxQexpectation}) suggests that it might be advantageous to reduce the ability of the critic to learn the high frequency components of the state-action value function, \(Q^{HIGH}\). Some of this already happens because the critic is going to learn first the low frequency components of Q. Our hypothesis is that we can increase learning efficiency by \emph{explicitly} reducing the importance of the high frequency components, and by emphasizing the low frequency components that have the most impact on the policy improvement. 

\subsection{Bandlimited Bellman residual}

In the classic SAC algorithm, we learn the parameters of the state-action value approximation by minimizing the mean squared Bellman residual,
\begin{equation} \label{criticlossrepeat}
J_Q(\theta) = E_{\boldsymbol{s}_t,\boldsymbol{a}_t}\left[ \frac{1}{2} \left( Q_{\theta}(\boldsymbol{s}_t,\boldsymbol{a}_t) - Q'(\boldsymbol{s}_t,\boldsymbol{a}_t)\right)^2 \right]
\end{equation} 
where 
\begin{equation} \label{Qprime}
Q'(\boldsymbol{s}_t,\boldsymbol{a}_t)=r(\boldsymbol{s}_t, \boldsymbol{a}_t)+E_{\boldsymbol{s}_{t+1}} \left[ V(\boldsymbol{s}_{t+1}) \right]
\end{equation} 
is the target state-action value function, and \(V(\boldsymbol{s}_{t+1})\) is the target value function, which itself is learnt by a third nework that minimizes the Bellman residual for the value approximation. In the bandlimited SAC, we minimize instead a \emph{bandlimited} version of the Bellman residual,
\begin{equation} \label{criticlossLOW}
J_Q(\theta) = E_{\boldsymbol{s}_t,\boldsymbol{a}_t}\left[ \frac{1}{2} \left( Q_{\theta}(\boldsymbol{s}_t,\boldsymbol{a}_t) - Q'^{LOW}(\boldsymbol{s}_t,\boldsymbol{a}_t)\right)^2 \right]
\end{equation} 
\(Q'^{LOW}\), in turn, can be computed using the Bellman operator modified with the addition of a bandlimiting filter operation \(h^{LOW}\) on the next-state prediction of the target critic: 
\begin{equation} \label{QprimeLOW}
Q'^{LOW}(\boldsymbol{s}_t,\boldsymbol{a}_t) =
r(\boldsymbol{s}_t, \boldsymbol{a}_t) + 
E_{\boldsymbol{s}_{t+1},\boldsymbol{a}_{t+1}} \left[ \left( h^{LOW}*Q_{\theta'}(\boldsymbol{s}_{t+1},.) \right) (\boldsymbol{a}_{t+1}) \right]
\end{equation} 
Note that Eq. (\ref{QprimeLOW}) does not rely on a value network to approximate the next-state value as in classic SAC; it uses instead a target critic network whose parameters are updated periodically to match the current critic parameters. Omitting the value network may increase the variance of the learnt state-action value approximation and possibly reduce sample efficiency.

\subsection{Closed form solution with linear approximation}

The classic SAC admits a closed form solution for the particular case of linear state-action value approximation in a discrete action and state spaces [14]. In this section we'll use \(\boldsymbol{q_{\omega}}\) to refer to the Q function to distinguish it from the non linear case, we'll use \(\boldsymbol{\omega}\) to refer to the linear parameters and \(\boldsymbol{\Phi}\) to the feature matrix,
\begin{equation} \label{Qw_main}
\boldsymbol{q_{\omega}} = \boldsymbol{\Phi} \boldsymbol{\omega}
\end{equation}
The bandlimited SAC also admits a closed form solution that can be derived in a similar fashion as in [14], as we show in the appendix C. The solution exploits the orthogonality of the convolutional filter with the high frequency Fourier basis functions in the action space. To see this, it is convenient to define the feature matrix as 
\begin{equation} \label{PhiFourierLowAndHigh_main}
\boldsymbol{\Phi} =
\begin{bmatrix}
\boldsymbol{\phi^{L}} & \boldsymbol{0} & \boldsymbol{0} & ... & \boldsymbol{\phi^{H}} & \boldsymbol{0} & \boldsymbol{0} & ...\\
\boldsymbol{0} & \boldsymbol{\phi^{L}} & \boldsymbol{0} & ... & \boldsymbol{0} & \boldsymbol{\phi^{H}} & \boldsymbol{0} & ...\\
\boldsymbol{0} & \boldsymbol{0} & \boldsymbol{\phi^{L}} & ... & \boldsymbol{0} & \boldsymbol{0} & \boldsymbol{\phi^{H}} & ...\\
... & ... & ... & ... & ... & ... & ... & ...
\end{bmatrix}
\end{equation}
where \(\boldsymbol{\phi^{L}}\) and \(\boldsymbol{\phi^{H}}\) are formed by column vectors that are themselves low and high frequency elements of the discrete Fourier basis in the action space, and are stacked along the diagonal \( \mathcal{S} \) times, one for each discrete state. The convolutional filter is orthogonal to each of the column vectors in \(\boldsymbol{\phi^{H}}\), and the state-action target critic approximation for bandlimited SAC retains \emph{only} the low frequency terms,
\begin{equation} \label{QwLowAndHighBL_main}
\boldsymbol{q_{\omega}}^{LOW} = 
\begin{bmatrix}
\boldsymbol{\Phi^{L}} & \boldsymbol{\Phi^{H}}
\end{bmatrix}
\begin{bmatrix}
\boldsymbol{\omega^{L}} \\
\boldsymbol{0} \\
\end{bmatrix} =
\boldsymbol{\Phi^{L}} \boldsymbol{\omega^{L}}
\end{equation}
We can now write the expression for the projected Bellman error for the bandlimited SAC as, 
\begin{equation} \label{eq15original_main}
\boldsymbol{\Phi^{T}} N_{\pi}
\left( 
\boldsymbol{r} +  \gamma \boldsymbol{P_{\pi}} ( \boldsymbol{q_{\omega}}^{LOW} - \boldsymbol{\Omega_{\pi}} ) - \boldsymbol{q_{\omega}} 
\right) = 0
\end{equation}
where \(N_{\pi}\) is the visitation density matrix, \(\boldsymbol{P_{\pi}}\) is the state transition matrix, and \(\boldsymbol{\Omega_{\pi}}\) the entropy regularization term. We show in the appendix C that the solution for \(\boldsymbol{\omega^{L}}\) can be obtained by adding a correction term to the \emph{low resolution solution} (the solution that uses \emph{only} low frequency basis functions),
\begin{equation} \label{WL_1_OLS_main}
\boldsymbol{\omega^{L*}}
=
\boldsymbol{\omega^{LowRes,L*}} - 
\boldsymbol{\Gamma^{LH}} \boldsymbol{\omega^{H*}}
\end{equation}
One way to interpret the bandlimited SAC fixed-point solution is to think of the low resolution as an starting point for the steady state solution. The bandlimited SAC critic learns first the low resolution fixed-point solution, and continues then 'adding' the effects of the high frequency terms until the projected Bellman error becomes zero. Each of the high frequency terms learnt (each of the components of \(\boldsymbol{\omega^{H*}}\)) affects the values of all the low frequency coefficients because of the presence of the coupling matrix \(\boldsymbol{\Gamma^{LH}}\) in eq (\ref{WL_1_OLS_main}). We show in the appendix C that, whereas in the classic SAC the matrix \(\boldsymbol{\Gamma^{LH}}\) is densely populated and each high frequency term influences multiple low frequency terms simultaneously, in the bandlimited SAC the effect is localized because \(\boldsymbol{\Gamma^{LH}}\) is a off-diagonal block of a matrix that becomes identically null for high entropy policies, something that doesn't happen in classic SAC. Although the rigorous characterization of \(\boldsymbol{\Gamma^{LH}}\) in the classic and bandlimited cases with moderate or zero temperature values is something that we are currently investigating, the limit behavior in the case of uniform policy suggests that the bandlimiting filter is able to reduce the influence of the high frequency terms on the low frequency terms, maybe allowing the critic to learn the solution coefficients faster than in classic SAC.

\begin{figure}[t]
%   \centering
% %   \fbox{\rule[-.5cm]{0cm}{4cm} \rule[-.5cm]{4cm}{0cm}}
  \includegraphics[width=14cm, height=9cm]{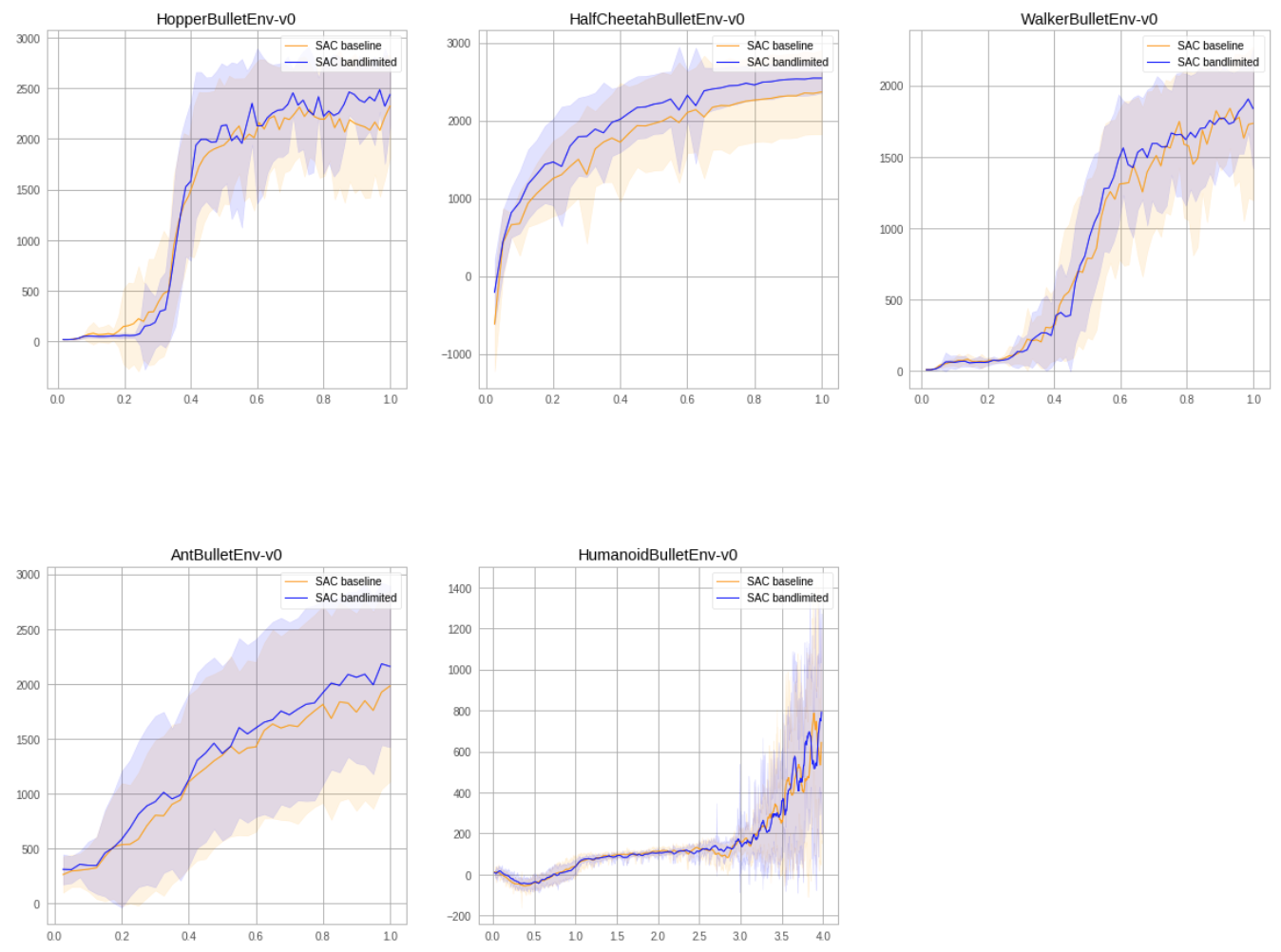}
  \caption{Performance of SAC-bandlimited (blue) against SAC-baseline (orange) across 5 PyBullet tasks (higher is better). Horizontal axis represents number of steps (million). Vertical axis represents episode reward.}
\end{figure}

\section{A filter that adapts to temperature values}

An approach used in practice to smooth out the the value estimate is to fit the value of a small region of the action space around the target action. The expectation over nearby actions is typically approximated by adding a small random perturbation to the target policy without the need to explicitly perform the expectation operation because the algorithm is already averaging over the mini-batches. In contrast, our proposed architecture explicitly performs a functional integration of the state-action value approximation over nearby action values.  Specifically, we bootstrap the output of the target critic(s) and feed it as input to a convolutional layer. A dense convolutional layer would require \(O(n^{N_A})\) additional operations, with \(n\) being the size of the convolutional kernel and \(N_A\) the dimensionality of the action space. This would be very costly and hard to implement with large action spaces. We opt instead to implement a separable convolutional filter that has non zero kernel weights only along the main axes of the action space.  This reduces the computational burden of the filter down to \(O(nN_A)\), which is more practical. The filtered value of the target critic is then:
\begin{equation} \label{criticlossLOW}
\left( h^{LOW}*Q_{\theta'}(\boldsymbol{s}_{t+1},.) \right) (\boldsymbol{a}_{t+1}) = \\
\sum_{1\leq i\leq N_A} \sum_{-K\leq k\leq K} \alpha_{k}
Q_{\theta'}\left(\boldsymbol{s}_{t+1},\boldsymbol{a}_{t+1} + \delta_k \boldsymbol{e}_i
\right)
\end{equation} 
where \(\boldsymbol{e}_i\) represents the unitary vectors along each of the dimensions of the action space. \(\alpha_{k}\) and \(\delta_k\) represent the filter weights and the location of the action coordinate shifts along the main dimensions, respectively,
\begin{equation} \label{finitefilter}
\alpha_{k} = \frac {sin \left( \pi x_k \right) }{ \pi x_k}
\end{equation} 
\begin{equation} \label{deltas}
\delta_k = x_k \frac {\pi}{w_{cutoff}}
\end{equation} 
with \(x_k = k/K\), \(-K\leq k\leq K\) and \(w_{cutoff}\) as given by eq (\ref{main_cutoff}). The specific choice of weights in eq. (\ref{finitefilter}) is convenient because their frequency response (the inverse Fourier transform) is a perfect low pass filter. Other choices are possible though. Eq (\ref{deltas}) captures the shift in the action values used to sample the Q approximation. The scaling factor \(\pi / w_{cutoff} \) determines the spatial resolution of the filter. The filter defined by eqs (\ref{finitefilter}) and (\ref{deltas}) has the frequency cutoff at frequencies close to \(w_{cutoff}\), which means that it filters out spatial components of the state-action value function with frequencies above \(w_{cutoff}\). Since \(w_{cutoff}\) in eq (\ref{main_cutoff}) is the inverse of the policy standard deviation, the filter will have high spacial resolution (higher \(w_{cutoff}\)) when the actor network chooses low exploration policies, and vice versa. In that sense, we say that the filter adapts to temperature values.

In practice, there are two main aspects in which the filter deviates from an idealized low-pass filter with cutoff frequency at \(w_{cutoff}\). First, the size of the filter, K, must be kept small to avoid high computational cost. In all the experiments we report below, we kept K=2,3 or 4 (filter length 5, 7 or 9). Second, it's necessary to bind action values between (-1,1). To do that, we apply the shift \(\delta_k \boldsymbol{e}_i\) not on the action itself, but on the inverse logistic transform of the action. Finally, a key difference with convolutional filters used in other deep learning applications is that the weights of the filter are fixed, not learnt. In future versions we plan to experiment with trainable filters. To reduce overestimation bias, we use a twin-critic approach as described in [1]. In the twin critics architecture, we train two independent critics and select the minimum of the two target critics to compute the Bellman residual and update the two critic networks. Our proposed architecture does not use a value network and this may result in high variance in the learnt state-action value approximation. To reduce variance, we use the twin critics also during policy updates: whereas classic SAC implementations use only one of the two critics during the policy update, we use the mean value of the two critics. Finally, the temperature parameter is optimized automatically following [1] unless indicated otherwise.

\begin{table}[t]
\caption{AVERAGE \textbf{MAX} Return, SAC SOTA vs Bandlimited SAC, 30 trials 1M steps each (for Humanoid 10 trials 4M steps). SOTA results are based on MuJoCo simulator (except Humanoid, for which SOTA is based on PyBullet).}
\label{table:SOTA_vs_bandlimited}
\vskip 0.15in
\begin{center}
\begin{small}
\begin{sc}
\begin{tabular}{lcccr}
\toprule
Environment & SOTA & Bandlimited\\
\toprule
HopperBulletEnv-v0        &  \textbf{3000} &  2645$\pm$  128  \\
HalfCheetahBulletEnv-v0   &  2350 &  \textbf{2581}$\pm$  208  \\
WalkerBulletEnv-v0        &  1285 &  \textbf{1979}$\pm$  219  \\
AntBulletEnv-v0           &   655 &  \textbf{2255}$\pm$  705  \\
HumanoidBulletEnv-v0      &  1265 &  \textbf{1443}$\pm$  172  \\
\bottomrule
\end{tabular}
\end{sc}
\end{small}
\end{center}
\vskip -0.1in
\end{table}

\begin{table}[t]
\caption{AVERAGE \textbf{MEAN} Return, SAC Baseline vs Bandlimited SAC, 30 trials 1M steps each (for Humanoid 10 trials 4M steps).}
\label{table:classic_vs_bandlimited}
\vskip 0.15in
\begin{center}
\begin{small}
\begin{sc}
\begin{tabular}{lcccr}
\toprule
Environment & SAC & SAC & \(\delta\) & t-stat\\
  & baseline  & bandlmtd\\
\toprule
HopperBulletEnv-v0        &          2170$\pm$  429 &  \textbf{2411$\pm$  163 }&   241 &  \textbf{2.87} \\
HalfCheetahBulletEnv-v0   &          2307$\pm$  538 &  \textbf{2512$\pm$  210 }&   205 &  \textbf{1.94} \\
WalkerBulletEnv-v0        &          1755$\pm$  319 &  1797$\pm$  229 &    42 &  0.58 \\
AntBulletEnv-v0           &          1826$\pm$  834 &  \textbf{2045$\pm$  719} &   218 &  \textbf{1.09} \\
HumanoidBulletEnv-v0      &           727$\pm$  332 &   794$\pm$  283 &    67 &  0.48 \\

\bottomrule
\end{tabular}
\end{sc}
\end{small}
\end{center}
\vskip -0.1in
\end{table}

\begin{table}[t]
\caption{Difference in population mean of simulations \textbf{MEAN} returns, and t-statistic, using \textbf{fixed-temperature} bandlimited SAC without and with noise. 30 trials 1M steps each (for Humanoid 10 trials 4M steps).}
\label{table:noiseless_vs_noisy}
\vskip 0.15in
\begin{center}
\begin{small}
\begin{sc}
\begin{tabular}{lcccr}
\toprule
Environment & G/L &  & G/L &  \\
  & w/o  & & w/\\
  & noise  & t-stat & noise & t-stat\\
\toprule
HopperBulletEnv-v0        &            77 &  0.7 &  \textbf{ 120 }&\textbf{ 1.4} \\
HalfCheetahBulletEnv-v0   &          \textbf{ -97 }&\textbf{ -1.1} &  \textbf{ 233 }&\textbf{ 2.0} \\
Walker2DBulletEnv-v0        &            -8 & -0.1 &    70 & 0.7 \\
AntBulletEnv-v0           &           144 &  0.7 &   \textbf{152 }&\textbf{ 1.1} \\
HumanoidBulletEnv-v0      &            46 &  0.3 &   \textbf{168 }&\textbf{ 1.9} \\
\bottomrule
\end{tabular}
\end{sc}
\end{small}
\end{center}
\vskip -0.1in
\end{table}

\section{Experiments}

We tested the convolutional filter in five OpenAI Gym environments running on the PyBullet physics simulator [16], which is used often in sim-to-real transfer tasks because of its realistic collision detection [4, 17]. Table \ref{table:SOTA_vs_bandlimited} shows the average max. returns of simulations for the proposed architecture, table \ref{table:classic_vs_bandlimited} isolates the effect of adding a bandlimiting filter to the baseline SAC, and table \ref{table:noiseless_vs_noisy} shows the effect of using stochastic rewards in a fixed temperature setting. The state of the art results are based on MuJoCo simulator (except Humanoid, for which SOTA is based on PyBullet). Environments are shown in order of increasing dimensionality of the action space. The choice of hyperparameters is shown in the appendix A. All environments use the same set of hyperparameters except Humanoid, which has the most complex action space (17 dimensions, vs 3 in Hopper, 6 in Half Cheetah and Walker, and 8 in Ant). For Humanoid simulations, we increased x4 the number of steps. Our proposed bandlimited SAC algorithm differs from classic SAC in two aspects: we don't use a value network to reduce value estimation variance, and we apply a bandlimiting filter at the output of the twin target critics. All \emph{baseline} simulations use automatic temperature tuning. \emph{Bandlimited} simulations use also automatic temperature tuning except Ant, for which we set temperature fixed (\(\alpha = 0.05\)).

Table \ref{table:SOTA_vs_bandlimited} shows average MAX returns of bandlimited SAC compared to SOTA classic SAC results. Bandlimited SAC outperforms SOTA SAC in three out of the five environments, suggesting that reducing the influence of high frequency terms on low frequency terms in the state-action value approximation can help increase sample efficiency. The exception is Hopper, which is the simplest of the five environments. In table \ref{table:classic_vs_bandlimited} we explore in more detail the effect of adding a bandlimiting filter to our baseline SAC implementation. For this analysis we show population average and standard deviation of simulation MEAN returns, as well as the difference in population means. From inspection of the table we can see that the standard deviation is in general smaller for bandlimited SAC, which means that bandlimited SAC returns are \emph{more consistent} than baseline SAC returns. To help compare the mean returns from the two populations with different standard deviations, we also show in the table the t-statistics of the difference in the population means. Figure 1 and table \ref{table:classic_vs_bandlimited} shows that the addition of the bandlimiting filter increases the average mean return in three out of the five cases (t-statistic greater than 1). Humanoid and Walker results are neutral.

The main hypotheses in this study is that the bandlimited SAC increases sample efficiency because it reduces the influence of high frequency components on the state-action value approximation. To illustrate this effect, we run a set of simulations in which we explicitly add stochastic noise in the reward observed by the agent, and compare the effect of omitting or adding the bandlimiting filter. For this comparison we fix the value of the temperature in both models to remove the effect of temperature optimization. We simulate noise in rewards using an additive noise process that is self-correlated in the temporal and the spatial (in the action space) domain. This means that it we take two samples of the noise process at two different but proximal spatial-temporal coordinates, the two samples will be similar. Self-correlated noise simulates imperfections in real world environments in which friction, inertia, turbulence, etc introduce correlated disturbances in the agent's observations. In the inverted pendulum, for instance, random horizontal forces acting on the pendulum mass caused by wind, friction, etc, introduce random disturbances in velocity and position [7], and thus in rewards. When wind or friction forces have some dependence with position, velocity or acceleration--in an unmanned aerial vehicle, for instance, air turbulence intensity and direction can change slowly with vertical position and with time [10]--the disturbances induced in the reward function themselves are self-correlated (see appendix D for an illustration). Importantly, self-correlated disturbances in the action space can't be \emph{completely} averaged out through Monte Carlo sampling in the action space, or with a replay buffer, and will eventually introduce disturbances in the Q function. In contrast, white (uncorrelated) noise will likely be averaged out and is less likely to introduce steady state disturbances in the Q function. To introduce temporal correlation, we use a Ornstein-Uhlenbeck process [8]. To introduce correlation in the action space, we discretize the action space using a finite hyperrectangular grid and simulate a different Ornstein-Uhlenbeck process for each point of the grid, which we update at the beginning of each episode [12]. Table \ref{table:noiseless_vs_noisy} shows that, in general, the addition of the bandlimiting filter with fixed temperature didn't increase returns in the noiseless case. Bandlimiting increased returns though in the presence of noise-in-rewards (even in the absence of temperature optimization). We think this is because self-correlated stochastic disturbances in the reward observed by the agent induce spurious high frequency terms in the state-action value approximation during early training. The bandlimiting filter is helpful because it allows the critic network to \emph{not pay too much attention} to the noisy terms.

\section{Conclusion}

A bandlimiting filter can help reduce the interdependency between the low frequency and the (harder to learn) high frequency components of the state-action value approximation and improve learning. Surprisingly, bandlimited SAC can have better asymptotic returns and stability than baseline SAC, suggesting that the high frequency components of the value approximation can be more difficult to learn. Analysis of the closed form solution suggests that the decoupling between the low and high frequency components is total for uniform policies, and vanishes as policies become deterministic. Temperature values in the bandlimited SAC algorithm have a dual role as they control the entropy of the optimal policy and the degree of interdependency between low and high frequency components of the value approximation. 

\section*{Broader Impact}

This work explores the properties of the value function in existing reinforcement learning algorithms from the optics of the frequency domain. The analysis leads to novel insights that can open new avenues of research in an area --frequency domain-- where most of the research has been done in the context of supervised learning. In addition, our analysis proves that we can capture improvements in training efficiency with simple modifications of existing algorithms, something that is important for reduced energy consumption. This work doesn't introduce new societal or ethics considerations or aggravate existing ones, and doesn't exploit new bias in the data.

\section*{References}

\medskip

\small

[1] Haarnoja, T., Zhou, A., Abbeel, P.\ \& Levine, S.\ (2018) Soft actor-critic: Off-policy maximum entropy deep reinforcement learning with a stochastic actor. {\it arXiv:1801.01290}.

[2] Nachum, O., Norouzi, M., Tucker, G.\ \& Schuurmans, D. \ (2018) Smoothed action value functions for learning gaussian policies. {\it arXiv:1803.02348}.

[3] Fellows, M., Ciosek, K\ \& Whiteson. S.\ (2018) Fourier policy gradients. {\it Proceedings of the 35th International Conference on Machine Learning}, ICML, Stockholmsmässan, Stockholm, Sweden.

[4] Mazoure, B., Doan, T., Durand, A., Hjelm, R. D.\ \& Pineau, J.\ (2019) Leveraging exploration in off-policy algorithms via normalizing flows.. {\it arXiv:1905.06893}.

[5] Rahaman, N., Baratin, A., Arpit, D., Draxler, F., Lin, M., Hamprecht, F., Bengio, Y.\ \& Courville, A.\ (2019) On the spectral bias of neural networks. {\it Proceedings of the 36th International Conference on Machine Learning}.

[6] Dziedzic, A., Paparrizos, J., Krishnan, S., Elmore, A.\ \& Franklin, M.\ (2019) Band-limited training and inference for convolutional neural networks. {\it International Conference on Machine Learning.}

[7] Prasad, L.B., Tyagi, B. \ \& Gupta, H.O.\ (2014) Optimal Control of Nonlinear Inverted Pendulum System Using PID Controller and LQR: Performance Analysis Without and With Disturbance Input. {\it International Journal of Automation and Computing}.

[8] Uhlenbeck, G.E \ \& Ornstein, L.S.\ (1930)  On the theory of the brownian motion. { \it Physical review.} Vol 36, No. 5.

[9] Tessler, C., Efroni, Y. \ \& Mannor S. \ (2019) Action Robust Reinforcement Learning and Applications in Continuous Control. {\it Proceedings of the 36th International Conference on Machine Learning}, Long Beach, California.

[10] Symington A., De Nardi R., Julier S. \ \& Hailes S. \ (2014) Simulating Quadrotor UAVs in Outdoor Scenarios. {\it IEEE/RSJ International Conference on Intelligent Robots and Systems}.

[11] Fujimoto, S., van Hoof, H., \ \& Meger, D. \ (2018) Addressing function approximation error in actor-critic methods. {\it arXiv preprint arXiv:1802.09477}.

[12] Peng, X.B., Andrychowicz, M., Zaremba, W. \ \& Abbeel P. \ (2017) Sim-to-real transfer of robotic control with dynamics randomization. {\it arXivpreprint arXiv:1710.06537}.

[13] Feng, Y., Li, L. \ \& Liu, Q.\ (2019) A Kernel Loss for Solving the Bellman Equation. {\it arXiv preprint arXiv:1905.10506}.

[14] Suttle, W., Yang, Z., Zhang, K., \ \& Liu, J. \ (2019) A Convergence Result for Regularized Actor-Critic Methods. {\it NeurIPS 2019}.

[15] Sutton, R. S. \ \& Barto, A. G. \ (2018) Reinforcement Learning, An Introduction. The MIT Press, 2nd.

[16] Coumans, E. \ \& Bai, Y. \ (2016) Pybullet, a python module for physics simulation for games, robotics and machine learning. GitHub repository.

[17] Duan, Y.,Chen, X.,Houthooft, R.,SchulmanJ. \ \& Abbeel, P. \ (2016) Benchmarking deep reinforcement learning for continuous control. {\it International Conference on Machine Learning, pages 1329– 1338}.

[18] R. Fox, A. Pakman \ \& N. Tishby \ (2016) Taming the noise in reinforcement learning via soft updates. {\it Conference on Uncertainty in Artificial Intelligence}.

[19] Xu, Z., Zhang, Y., Luo, T., Xiao, Y. \ \& Ma, Z. \ (2019) Frequency Principle: Fourier analysis sheds light on deep neural networks. {\it arXiv:1901.06523v3}.

[20] Bellemare, M. G., Dabney, W. \ \& Munos, R \ (2017). A distributional perspective on reinforcement learning. {\it International Conference on Machine Learning (ICML)}.

\section*{Appendix A. Code and hyperparameters}

We ran the simulations using ReAgent opensource RL software\footnote{https://github.com/facebookresearch/ReAgent} with the following parameters

\label{table:hyperpaams}
\vskip 0.15in
\begin{center}
\begin{small}
\begin{sc}
\begin{tabular}{lcccr}
\toprule
Hyperparameter & Value\\
\toprule
alpha      &    automatic tuning (except Ant, 0.05)\\
gamma      &    0.99 \\
tau      &    0.005 \\
learning rate      &    0.003 \\
batch size      &    256 \\
updates per step      &    1 \\
target update interval      &    1 \\
steps      &    1M (4M for Humanoid) \\
training starts      &    5,000 steps \\
replay size & 1M \\
filter length & 7 for Hopper, 9 for Half Cheetah, 5 for the rest \\
\bottomrule
\end{tabular}
\end{sc}
\end{small}
\end{center}
\vskip -0.1in

\section*{Appendix B. Expected Q in the frequency domain}

Using the classical Gaussian reparametrization with diagonal covariance matrix, it's easy to see that, conditional on \(\boldsymbol{s}_t\), the state-action value expectation term in the policy loss function is a convolution of the state-action value function with a Gaussian kernel evaluated at the mean policy value \(\mu_{\phi}(s_t)\),
\begin{multline} \label{becomesconv}
E_{\boldsymbol{s}_{t}\sim D,\boldsymbol{\epsilon}_{t}\sim N} \left(Q_{\theta}(\boldsymbol{s}_t,f_{\phi}(\boldsymbol{\epsilon}_{t},\boldsymbol{s}_t))
\right) \approx \\
E_{\boldsymbol{s}_{t}\sim D} \left[ 
\int Q_{\theta}\left(\boldsymbol{s}_t, \mu_{\phi}(\boldsymbol{s}_t) + \sigma_{\phi}(\boldsymbol{s}_t) \boldsymbol{\epsilon}_t\right)
\frac{1}{(2\pi)^\frac{n}{2}}e^{-\frac{1}{2}\boldsymbol{\epsilon}_t^T\boldsymbol{\epsilon}_t}d\boldsymbol{\epsilon}_t \right]
\\ = E_{\boldsymbol{s}_{t}\sim D}  
\left[
\left( 
Q_{\theta}(\boldsymbol{s}_t, .) *  
f_N( .;\sigma_{\phi}(\boldsymbol{s}_t) )
\right)
_{\mu_{\phi}(\boldsymbol{s}_t)}
\right]
\end{multline}
where \(\mu_{\phi}(\boldsymbol{s}_t)\) and \(\sigma_{\phi}(\boldsymbol{s}_t)\) are the mean column vector and diagonal covariance matrix, respectively, * is the convolution operator and \(f_N\) is the Gaussian probability distribution,
\begin{equation} \label{fN}
f_N(\boldsymbol{z};\sigma)=\frac{e^{-\frac{1}{2} 
\boldsymbol{z}^T \left[ \sigma^{T} \sigma \right]^{-1} \boldsymbol{z}
}}{|\sigma|(2\pi)^\frac{n}{2}}
\end{equation}
The reason eq. (\ref{becomesconv}) is only an approximation is because we ignore in this section the \(tanh\) transformation in the parametrized action. The importance of rewriting the action value expectation this way is that it admits also an interpretation in the dual frequency domain (the frequency domain of the action space). In effect, remembering that the Fourier transform of the convolution is equivalent to the product of Fourier transforms, we can then rewrite the state-action value expectation as
\begin{equation} \label{productofFs}
E_{\boldsymbol{s}_{t}\sim D,\boldsymbol{\epsilon}_{t}\sim N} \left(Q_{\theta}(\boldsymbol{s}_t,f_{\phi}(\boldsymbol{\epsilon}_{t},\boldsymbol{s}_t))
\right) \approx 
E_{\boldsymbol{s}_{t}\sim D}  
\left(
F^{-1}
\left[
F_{Q_{\theta}}  F_{f_{N}}
\right]
\left[ 
\mu_{\phi}(\boldsymbol{s}_t) 
\right]
\right)
\end{equation}
The first Fourier transform \(F_{Q_{\theta}}\) is the Fourier transform of the state-action value function.  The second Fourier transform \(F_{f_{N}}\) is the Fourier transform of the Gaussian probability distribution in the action space with standard deviation \( \sigma^{i}(\boldsymbol{s}_t) \), which is itself a Gaussian in the frequency domain with standard deviation \( \frac{\pi}{\sigma^{i}(\boldsymbol{s}_t)} \). This second Fourier transform can be seen therefore as a low pass filter in the frequency domain that dampens or eliminates the high frequency content of the Q function. Practically speaking, this filter eliminates frequency components in Q for frequencies above a cutoff value of
\begin{equation} \label{cutoff}
w^{i} > w_{cutoff}^{i} \approx \frac{\pi}{2\sigma_{\phi}^{i}(\boldsymbol{s}_t)}
\end{equation} 
In practical terms, this means that the high frequency content in the state-action value function has little or no impact on the state-action value expectation (it gets averaged out in the convolution) or on the overall policy loss, and we can thus write approximately
\begin{equation} \label{approxQexpectation}
E_{\boldsymbol{s}_{t}\sim D,\boldsymbol{\epsilon}_{t}\sim N} \left(Q_{\theta}(\boldsymbol{s}_t,f_{\phi}(\boldsymbol{\epsilon}_{t},\boldsymbol{s}_t))
\right) 
\approx
E_{\boldsymbol{s}_{t}\sim D,\boldsymbol{\epsilon}_{t}\sim N} \left(Q_{\theta}^{LOW}(\boldsymbol{s}_t,f_{\phi}(\boldsymbol{\epsilon}_{t},\boldsymbol{s}_t))
\right)
\end{equation}
where \(Q_{\theta}^{LOW}\) refers to the the function that results from removing high frequency components from the Fourier transform \(F_{Q_{\theta}}\),
\begin{equation} \label{Qlow}
Q_{\theta}^{LOW} = F^{-1} \left[
F_{Q_{\theta}} \boldsymbol{1_{w<\frac{\pi}{2\sigma_{\phi}(\boldsymbol{s}_t)}}}\right]
\end{equation}
For this study, we didn't directly filter out the high frequency components from the Q approximation in the critic network. Instead, we filtered in the output of the \emph{target} critic network. In future work we could explore direct filtering on the critic network itself.

\section*{Appendix C. Linear approximation}

Suttle et al. [14] offer a convergence proof for entropy-regularized Actor Critic algorithms in discrete action and state spaces and linear state-action value approximation. We explore a closed form solution for the bandlimited SAC using similar arguments as theirs.

Let \(\boldsymbol{q_{\pi}} \in \mathbb{R}_{|\mathcal{A}||\mathcal{S}|\times 1}\) be the vector of state-action values
\begin{equation} \label{Qvector}
\boldsymbol{q_{\pi}} = [ q_{\pi}(s_{1},a_{1}), q_{\pi}(s_{1},a_{2}), ..., q_{\pi}(s_{1},a_{|A|}), 
q_{\pi}(s_{2},a_{1}), ..., q_{\pi}(s_{|S|},a_{|A|}) ]^{T}
\end{equation}
and \(\boldsymbol{q_{\omega}} \) its linear approximation 
\begin{equation} \label{Qw}
\boldsymbol{q_{\omega}} = \boldsymbol{\Phi} \boldsymbol{\omega}
\end{equation}
where the columns in \( \boldsymbol{\Phi} \in \mathbb{R}_{|\mathcal{A}||\mathcal{S}|\times K}\) are all linearly independent, and \(K\ll |\mathcal{A}|\). \( \boldsymbol{\omega} \in \mathbb{R} _{K\times 1}\) is the column vector of coefficients that can be learnt with an iterative schema as we describe next. Let \( \boldsymbol{r} \in \mathbb{R} _{|\mathcal{A}||\mathcal{S}|\times 1}\) be the column vector of reward values; \( \boldsymbol{\Xi_{\pi}} \in \mathbb{R}_{|\mathcal{S}|\times 1}\) a column vector with policy entropy values, one per state, and \(\boldsymbol{\Omega_{\pi}} = \boldsymbol{\Xi_{\pi}} \otimes 1_{|\mathcal{A}|\times 1}\) the column vector of entropy values repeated \(|\mathcal{A}|\) times, \(\boldsymbol{\Omega_{\pi}} \in \mathbb{R}_{|\mathcal{A}||\mathcal{S}|\times 1}\). Let \( \boldsymbol{P_{\pi}} \in \mathbb{R}_{|\mathcal{A}||\mathcal{S}|\times |\mathcal{A}||\mathcal{S}|}\) be the action-state transition probability matrix. \( \boldsymbol{P_{\pi}}\) values are computed as \(p(s_{k}|s_{i},a{j})\pi(a_{l}|s_{k})\), which is the probability of transitioning from \((s_{i},a_{j})\) to \((s_{k},a_{l})\). The pairs \((s_{i},a_{j})\) and \((s_{k},a_{l})\) remain unchanged on a given row or column, respectively. Finally, let \( \boldsymbol{N_{\pi}} \in \mathbb{R} _{|\mathcal{A}||\mathcal{S}|\times 1}\) be the column vector with action-states visitation densities. With those definitions, the expression for the soft Bellman operator becomes

\begin{equation} \label{Tsoft}
\boldsymbol{T_{\pi}q} = \boldsymbol{r} +  \gamma \boldsymbol{P_{\pi}} ( \boldsymbol{q}-\boldsymbol{\Omega_{\pi}} )
\end{equation}
Suttle et al. [14] show that an iterative schema based on updating \( \boldsymbol{\omega} \) with the projected Bellman error converges to a fixed-point solution \(\boldsymbol{\omega^{*}}\) such that \(\boldsymbol{\Phi} \boldsymbol{\omega^{*}}\) is the fixed point of the projected Bellman error,
\begin{equation} \label{eq15original}
\boldsymbol{\Phi^{T}} N_{\pi}
\left( 
\boldsymbol{r} +  \gamma \boldsymbol{P_{\pi}} ( \boldsymbol{\Phi} \boldsymbol{\omega^{*}} - \boldsymbol{\Omega_{\pi}} ) - \boldsymbol{\Phi} \boldsymbol{\omega^{*}} 
\right) = 0
\end{equation}

\subsection*{Bandlimited target critic}
To study the bandlimited SAC algorithm, it is convenient to use the Fourier basis in the action space as the building block of the state-action value approximation, \(\boldsymbol{q_{\omega}}\). Let \(\left( \phi_{k}\in \mathbb{R} _{|\mathcal{A}|\times 1} \right), k=1...K \ll |\mathcal{A}|\) be the Fourier basis in the action space and \(\boldsymbol{\phi} \in \mathbb{R} _{|\mathcal{A}|\times K}\) the matrix with column vectors \(\phi_{1}, \phi_{2}, ...\) etc. We define the matrix \( \boldsymbol{\Phi} \) as
\begin{equation} \label{PhiFourier}
\boldsymbol{\Phi} =
\begin{bmatrix}
\boldsymbol{\phi} & \boldsymbol{0} & \boldsymbol{0} & ...\\
\boldsymbol{0} & \boldsymbol{\phi} & \boldsymbol{0} & ...\\
\boldsymbol{0} & \boldsymbol{0} & \boldsymbol{\phi} & ...\\
... & ... & ... & ...
\end{bmatrix}
\end{equation}
where we have repeated \( \boldsymbol{\phi} \) along the diagonal blocks \( |\mathcal{S}| \) times, one for each discrete state. The columns in \(\boldsymbol{\phi}\) can be rearranged into low frequency and high frequency columns, 
\begin{equation} \label{PhiLowAndHigh}
\boldsymbol{\phi} =
\begin{bmatrix}
\boldsymbol{\phi^{L}} & \boldsymbol{\phi^{H}}
\end{bmatrix}
\end{equation}
allowing us to also reorder the columns in eq. (\ref{PhiFourier}) and rewrite \( \boldsymbol{\Phi} \) as
\begin{equation} \label{PhiFourierLowAndHigh}
\boldsymbol{\Phi} =
\begin{bmatrix}
\boldsymbol{\phi^{L}} & \boldsymbol{0} & \boldsymbol{0} & ... & \boldsymbol{\phi^{H}} & \boldsymbol{0} & \boldsymbol{0} & ...\\
\boldsymbol{0} & \boldsymbol{\phi^{L}} & \boldsymbol{0} & ... & \boldsymbol{0} & \boldsymbol{\phi^{H}} & \boldsymbol{0} & ...\\
\boldsymbol{0} & \boldsymbol{0} & \boldsymbol{\phi^{L}} & ... & \boldsymbol{0} & \boldsymbol{0} & \boldsymbol{\phi^{H}} & ...\\
... & ... & ... & ... & ... & ... & ... & ...
\end{bmatrix}
\end{equation}
and to rewrite the expression for the linear approximation of the state-action value function, eq. (\ref{Qw}), as
\begin{equation} \label{QwLowAndHigh}
\boldsymbol{q_{\omega}} = 
\begin{bmatrix}
\boldsymbol{\Phi^{L}} & \boldsymbol{\Phi^{H}}
\end{bmatrix}
\begin{bmatrix}
\boldsymbol{\omega^{L}} \\
\boldsymbol{\omega^{H}} \\
\end{bmatrix} 
\end{equation}
\(\boldsymbol{\omega^{L}} \in \mathbb{R} _{K_{L}|\mathcal{S}|\times 1}\) and \(\boldsymbol{\omega^{H}} \in \mathbb{R} _{K_{H}|\mathcal{S}|\times 1}\) are the low and high frequency coefficients of the Fourier approximation. Because of the orthogonality of the convolutional filter with the high frequency components of the Fourier basis, the expression for the (filtered) target state-action value approximation is simply
\begin{equation} \label{QwLowAndHighBL}
\boldsymbol{q_{\omega}}^{target,LOW} = 
\begin{bmatrix}
\boldsymbol{\Phi^{L}} & \boldsymbol{\Phi^{H}}
\end{bmatrix}
\begin{bmatrix}
\boldsymbol{\omega^{L}} \\
\boldsymbol{0} \\
\end{bmatrix} =
\boldsymbol{\Phi^{L}} \boldsymbol{\omega^{L}}
\end{equation}
Substituting eq. (\ref{QwLowAndHighBL}) in eq. (\ref{eq15original}) and rearranging we have the expression of the projected Bellman error for the bandlimited state-action value approximation, 
\begin{equation} \label{PBE_bandlimited}
\boldsymbol{\Phi^{T}} N_{\pi} \boldsymbol{r} - \gamma \boldsymbol{\Phi^{T}} N_{\pi}\boldsymbol{P_{\pi}} \boldsymbol{\Omega_{\pi}}  
+ \gamma \boldsymbol{\Phi^{T}} N_{\pi}\boldsymbol{P_{\pi}} 
\boldsymbol{\Phi^{L}} \boldsymbol{\omega^{L*}} 
- \boldsymbol{\Phi^{T}} N_{\pi} \boldsymbol{\Phi} \boldsymbol{\omega^{*}} 
= 0
\end{equation}
The first and second terms represent, respectively, the Fourier coefficients of the (visitation density normalized) reward function and expected entropy value
\begin{equation} \label{Freward}
\boldsymbol{\Phi^{T}} N_{\pi} \boldsymbol{r} = 
\begin{bmatrix}
\boldsymbol{r^{\mathcal{F},L}} \\
\boldsymbol{r^{\mathcal{F},H}} \\
\end{bmatrix}
\end{equation}
\begin{equation} \label{FPOmega}
\boldsymbol{\Phi^{T}} N_{\pi}\boldsymbol{P_{\pi}} \boldsymbol{\Omega_{\pi}} = 
\begin{bmatrix}
\boldsymbol{\Omega^{\mathcal{F},L}} \\
\boldsymbol{\Omega^{\mathcal{F},H}} \\
\end{bmatrix}
\end{equation}
It helps if we define the soft reward Fourier coefficients as
\begin{equation} \label{SoftRewards}
\begin{bmatrix}
\boldsymbol{r^{soft,\mathcal{F},L}} \\
\boldsymbol{r^{soft,\mathcal{F},H}} \\
\end{bmatrix}
 = 
\begin{bmatrix}
\boldsymbol{r^{\mathcal{F},L}} \\
\boldsymbol{r^{\mathcal{F},H}} \\
\end{bmatrix}
- \gamma
\begin{bmatrix}
\boldsymbol{\Omega^{\mathcal{F},L}} \\
\boldsymbol{\Omega^{\mathcal{F},H}} \\
\end{bmatrix}
\end{equation}
and the \(\boldsymbol{U}\) and \(\boldsymbol{V}\) matrixes as
\begin{equation} \label{matrixU}
\gamma \boldsymbol{\Phi^{T}} N_{\pi}\boldsymbol{P_{\pi}} \boldsymbol{\Phi} =
\begin{bmatrix}
\boldsymbol{U^{LL}} & \boldsymbol{U^{LH}}\\
\boldsymbol{U^{HL}} & \boldsymbol{U^{HH}} \\
\end{bmatrix}
\end{equation}
\begin{equation} \label{matrixV}
\boldsymbol{\Phi^{T}} N_{\pi} \boldsymbol{\Phi} =
\begin{bmatrix}
\boldsymbol{V^{LL}} & \boldsymbol{V^{LH}} \\
\boldsymbol{V^{HL}} & \boldsymbol{V^{HH}} \\
\end{bmatrix}
\end{equation}
The projected Bellman error for a bandlimited SAC critic in eq. (\ref{PBE_bandlimited}) can now be written as
\begin{equation} \label{PBE_bandlimited_2}
\begin{bmatrix}
\boldsymbol{r^{soft,\mathcal{F},L}} \\
\boldsymbol{r^{soft,\mathcal{F},H}} \\
\end{bmatrix}
+
\begin{bmatrix}
\boldsymbol{U^{LL}} \boldsymbol{\omega^{L}}\\
\boldsymbol{U^{HL}} \boldsymbol{\omega^{L}}\\
\end{bmatrix} 
-
\begin{bmatrix}
\boldsymbol{V^{LL}} \boldsymbol{\omega^{L}} + \boldsymbol{V^{LH}} \boldsymbol{\omega^{H}} \\
\boldsymbol{V^{HL}} \boldsymbol{\omega^{L}} + \boldsymbol{V^{HH}} \boldsymbol{\omega^{H}}\\
\end{bmatrix}
= 0
\end{equation}
and, after grouping the terms that multiply the low and high frequency Fourier coefficients, we can write two matrix equations as
\begin{equation} \label{PBE_bandlimited_3}
\begin{matrix}
\boldsymbol{r^{soft,\mathcal{F},L}} 
+
( \boldsymbol{U^{LL}} - \boldsymbol{V^{LL}} )
\boldsymbol{\omega^{L}}
-
\boldsymbol{V^{LH}} \boldsymbol{\omega^{H}}
= 0 \\
\boldsymbol{r^{soft,\mathcal{F},H}} 
+
( \boldsymbol{U^{HL}} - \boldsymbol{V^{HL}} )
\boldsymbol{\omega^{L}}
-
\boldsymbol{V^{HH}} \boldsymbol{\omega^{H}}
= 0
\end{matrix} 
\end{equation}
We define the matrix \(\boldsymbol{\Delta} = \boldsymbol{V} - \boldsymbol{U}\) as follows,
\begin{equation} \label{DELTA}
\begin{bmatrix}
\boldsymbol{\Delta^{LL}} & \boldsymbol{\Delta^{LH}} \\
\boldsymbol{\Delta^{HL}} & \boldsymbol{\Delta^{HH}}\\
\end{bmatrix} = 
\begin{bmatrix}
\boldsymbol{V^{LL}} - \boldsymbol{U^{LL}} & \boldsymbol{V^{LH}} - \boldsymbol{U^{LH}} \\
\boldsymbol{V^{HL}} - \boldsymbol{U^{HL}} & \boldsymbol{V^{HH}} - \boldsymbol{U^{HH}}\\
\end{bmatrix}
\end{equation}
\(\boldsymbol{\Delta^{LL}}\) is equal to \( \boldsymbol{(\Phi^{L})^{T}} N_{\pi}(I - \gamma \boldsymbol{P_{\pi}}) \boldsymbol{\Phi^{L}}\) and is positive definite and invertible because it is a block matrix of \( \boldsymbol{\Phi^{T}} N_{\pi}(I - \gamma \boldsymbol{P_{\pi}}) \boldsymbol{\Phi}\), which is itself positive definite and invertible (see [15], p.206-207). The same argument applies to \(\boldsymbol{\Delta^{HH}}\). We can now write the low frequency part of the fixed point solution for the bandlimited SAC from the first eq. in (\ref{PBE_bandlimited_3}) as
\begin{equation} \label{WL_1}
\boldsymbol{\omega^{bl,L*}}
=
\boldsymbol{\Delta^{LL}}^{-1}
(\boldsymbol{r^{soft,\mathcal{F},L}} - \boldsymbol{V^{LH}} \boldsymbol{\omega^{bl,H*}})
\end{equation}
whereas for a classic SAC the expression is,
\begin{equation} \label{WL_1_classic}
\boldsymbol{\omega^{classic,L*}}
=
\boldsymbol{\Delta^{LL}}^{-1}
(\boldsymbol{r^{soft,\mathcal{F},L}} - \boldsymbol{\Delta^{LH}} \boldsymbol{\omega^{classic,H*}})
\end{equation}
The high frequency part of the fixed point solution, \(\boldsymbol{\omega^{bl,H*}}\), can be calculated from the second eq. in (\ref{PBE_bandlimited_3}) upon substituting the expression for \(\boldsymbol{\omega^{bl,L*}}\), leading to
\begin{equation} \label{WH_2}
\boldsymbol{\omega^{bl,H*}}
= 
(\boldsymbol{V^{HH}}
- \boldsymbol{\Delta^{HL}}
\boldsymbol{\Delta^{LL}}^{-1}
\boldsymbol{V^{LH}})^{-1}
(
\boldsymbol{r^{soft,\mathcal{F},H}} 
- \boldsymbol{\Delta^{HL}}\boldsymbol{\Delta^{LL}}^{-1}
\boldsymbol{r^{soft,\mathcal{F},L}}
)
\end{equation}
and, similarly for the classic SAC
\begin{equation} \label{WH_2_classic}
\boldsymbol{\omega^{classic,H*}}
= 
(\boldsymbol{\Delta^{HH}}
- \boldsymbol{\Delta^{HL}}
\boldsymbol{\Delta^{LL}}^{-1}
\boldsymbol{\Delta^{LH}})^{-1}
(
\boldsymbol{r^{soft,\mathcal{F},H}} 
- \boldsymbol{\Delta^{HL}}\boldsymbol{\Delta^{LL}}^{-1}
\boldsymbol{r^{soft,\mathcal{F},L}}
)
\end{equation}
\subsection*{Interpretation of the bandlimited SAC solution}
The orthogonality properties of the Fourier basis do not apply in SAC algorithms because of the presence of the diagonal visitation density matrix. We can however still rewrite the left hand side of eqs. (\ref{matrixV}) as 
\begin{equation} \label{matrixVSplitN}
\boldsymbol{\Phi^{T}} N_{\pi} \boldsymbol{\Phi} = (N_{\pi}^{1/2} \boldsymbol{\Phi} )^{T} (N_{\pi}^{1/2} \boldsymbol{\Phi} )
\end{equation}
and eq. (\ref{WL_1}) as
\begin{equation} \label{WL_1_OLS}
\boldsymbol{\omega^{bl,L*}}
=
\boldsymbol{\omega^{LowRes,L*}} - 
\boldsymbol{\Delta^{LL}}^{-1} 
(N_{\pi}^{1/2} \boldsymbol{\Phi^{L}} )^{T} (N_{\pi}^{1/2} \boldsymbol{\Phi^{H}} ) \boldsymbol{\omega^{bl,H*}}
\end{equation}
\(\boldsymbol{\omega^{LowRes,L*}} = \boldsymbol{\Delta^{LL}}^{-1} \boldsymbol{r^{soft,\mathcal{F},L}}\) is the fixed point solution when we \emph{only use low frequency (low resolution) components to approximate the state-action value function}. The bandlimited SAC solution is therefore equal to the low resolution SAC solution only in those cases in which the second term is null, which happens only for uniform policy parametrizations. In this situation, \((N_{\pi}^{1/2} \boldsymbol{\Phi^{L}} )^{T} (N_{\pi}^{1/2} \boldsymbol{\Phi^{H}} )=0\) because of orthogonality of the low and high frequency components of the Fourier basis. 

\subsection*{Noise in rewards (informal argument)}
The effect of self-correlated noise on the SAC fixed point solution can be modeled as a state-action dependent error affecting the Fourier coefficients of the reward function
\begin{equation} \label{NoiseinrewardsL}
\boldsymbol{r^{\mathcal{F},L}} \rightarrow \boldsymbol{r^{\mathcal{F},L}} + \boldsymbol{\epsilon^{L}}
\end{equation}
\begin{equation} \label{NoiseinrewardsH}
\boldsymbol{r^{\mathcal{F},H}} \rightarrow \boldsymbol{r^{\mathcal{F},H}} + \boldsymbol{\epsilon^{H}}
\end{equation}
Note that the added noise has different effects on the low frequency components of the fixed point solutions for the bandlimited and classic SAC algorithms. For the bandlimited SAC, the shift in fixed-point coefficients can be written as
\begin{equation} \label{errorinwl_bl}
\Delta \boldsymbol{\omega^{bl,L*}}
=
\boldsymbol{\Delta^{LL}}^{-1} \left( \boldsymbol{\epsilon^{L}} - 
\boldsymbol{V^{LH}} \boldsymbol{\Delta \omega^{bl,H*}} \right)
\end{equation}
whereas for classic SAC
\begin{equation} \label{errorinwl_classic}
\Delta \boldsymbol{\omega^{classic,L*}}
=
\boldsymbol{\Delta^{LL}}^{-1} \left( \boldsymbol{\epsilon^{L}} - 
\boldsymbol{\Delta^{LH}} \boldsymbol{\Delta \omega^{classic,H*}} \right)
\end{equation}
In the limit of uniform policy distribution, the matrix \(\boldsymbol{V^{LH}}\) is identically null because of orthogonality of the Fourier basis, whereas the matrix \(\boldsymbol{\Delta^{LH}}\) isn't. As the policy departs more and more from uniform, none of the two matrices is null, but the first one will attenuate the effect of the high frequency components if we choose \(\boldsymbol{\phi^{L}}\) and \(\boldsymbol{\phi^{H}}\) such that all the elements in \(\boldsymbol{\phi^{H}}\) have frequencies above the visitation density maximum frequency. When this occurs, the elements in \(\boldsymbol{\phi^{H}}\) oscillate multiple times within the interval of action values with positive density visitation values, whereas the elements of \(\boldsymbol{\phi^{L}}\) change only slowly within that interval, and this is likely to result in the following inequality  
\begin{equation} \label{attenuation}
|\boldsymbol{V^{LH}}| < |\boldsymbol{\Delta^{LH}}| 
\end{equation}
It is in that sense that we hypothesize that the bandlimited SAC helps eliminates high frequency noise from the low frequency components of the fixed point solution: it reduces the dependency of the low frequency components with the high frequency components.

\section*{Appendix D. Self-correlated noise for an inverted pendulum}

Prasad et al [7] provide formal derivation of state dynamics in an inverted pendulum with horizontal disturbances acting on the pendulum mass.  Near the equilibrium position, the angular acceleration of the pendulum mass and the horizontal acceleration of the cart can be written, respectively, as
\begin{equation} \label{AngularAcceleration}
\ddot \theta = Z_{1} \theta + Z_{2} u + Z_{3} \epsilon_{u}
\end{equation} 
\begin{equation} \label{LinearAcceleration}
\ddot x = Z_{4} \theta + Z_{5} u 
\end{equation} 
where \(\theta\) is the angle of the pendulum, \(u\) is the applied control, \(\epsilon_{u}\) is a random disturbance in the acceleration applied to the pendulum mass that depends on the magnitude and direction of the applied control and that models the influence of mechanical friction and other imperfections. \(Z_{1}, Z_{2}, ...\) are constants that depend on the physical properties of the system. Given a quadratic reward function like
\begin{equation} \label{PendulumReward}
r(u) = \alpha (\theta)^{2} + \beta (x)^{2}
\end{equation} 
we can show that, for the pair of actions \(u\) and \(u+\delta u\), if the random disturbances are correlated then the rewards observed by the agent are themselves correlated. In effect, after a small interval of time \(\delta t\) we can write the noisy rewards as
\begin{equation} \label{RewardNoisy} 
r^{noise}(u) \approx r(u) + 2 \alpha Z_{3} \frac{(\delta t)^{2}}{2} \epsilon_{u}
\end{equation} 
\begin{equation} \label{RewardNoisyDU}
r^{noise}(u+\delta u) \approx r(u+\delta u) + 2 \alpha Z_{3} \frac{(\delta t)^{2}}{2} \epsilon_{u+\delta u}
\end{equation} 
and their covariance as
\begin{equation} \label{RewardsCovariance}
cov(r(u),r(u+\delta u)) \approx \left( 2 \alpha Z_{3} \frac{(\delta t)^{2}}{2} \right)^{2} cov(\epsilon_{u},\epsilon_{u+\delta u})
\end{equation} 
Eq. (\ref{RewardsCovariance}) indicates that nonzero covariance in the random disturbance, \(cov(\epsilon_{u},\epsilon_{u+\delta u})\), can induce a nonzero covariance in the reward seen by the agent.

\end{document}